\documentclass{article}
\usepackage[final]{neurips_2026}
\makeatletter
\renewcommand{\@noticestring}{Workshop on Who Verifies the Agents? Toward Reliable Agent Development, NeurIPS 2026.}
\makeatother
\setcitestyle{numbers,square,comma}
\usepackage[utf8]{inputenc}
\usepackage[T1]{fontenc}
\usepackage{amsmath}
\usepackage{tabularx}
\usepackage{booktabs}
\usepackage{hyperref}
\usepackage{url}
\title{When Tool Calls Succeed but Workflows Fail:\\ Anomalies at the Agent--Tool Boundary}

\author{%
  Artem Trofimov \\
  AVIV Group \\
  Berlin, Germany \\
  \texttt{artemtrofimov@acm.org} \\
  \And
  Boris Novikov \\
  Independent Researcher \\
  Helsinki, Finland \\
  \texttt{borisnov@acm.org} \\
}

\begin{document}
\maketitle

\begin{abstract}
AI agents increasingly execute long-running workflows that externalize
effects through independently supplied tools. Under retries, speculative
execution, concurrency, and partial failures, the resulting external state
may be inconsistent with the workflow's intended resolution: required effects
may be missing or duplicated, aborted effects may survive, and committed
effects may depend on provisional state that is later withdrawn. Advanced
transaction models address related failures, but assume that
lower-level operations expose the semantics they depend on: whether an
effect occurred, whether it can be compensated, staged, or safely
reordered. Shared agent--tool interfaces usually do not. We contribute an effect-history model that separates events in the
external world from the runtime's observations of them, and a catalog of
eight recurring external-effect anomalies. From the catalog we derive the boundary
capabilities required to exclude each anomaly in general, and four points
where black-box tool invocation alone cannot provide a general guarantee. We then ask how much of this is expressible in a widely used shared tool
interface, measuring the use of the standard annotation vocabulary across
98{,}291 tools exposed by reachable remote servers in the official Model
Context Protocol (MCP) registry.
The fields are widely emitted but provide only coarse call-level hints, and
none of the required capabilities is fully expressible. These results motivate
reusable transactional contracts at the tool boundary.
\end{abstract}

\section{Introduction}

AI agents increasingly execute multi-step workflows whose operations
change the external world: they send messages, charge credit cards, place
orders, and invoke independently operated services. The correctness question
we study is not whether each tool call succeeds locally, but whether the
external effects that survive are consistent with a coherent resolution of
the workflow. Under retries, speculative execution,
concurrency, and partial failures, they need not be: one logical action may
occur twice, a workflow may commit without a required effect, an aborted
branch may leave residue, or a committed action may depend on an effect
that is later withdrawn.

This is a verification problem before it is a recovery problem: whether an
agent's execution can be checked at all (by the runtime itself, a supervising agent, or an external auditor) depends on what the tool
boundary makes observable. A verifier cannot establish that an effect
happened exactly once, survived an abort, or leaked to an outside observer
if the interface exposes no evidence either way.

These failures constitute a single consistency problem at the agent--tool
boundary. External effects
generally cannot be rolled back, may become visible before the workflow
resolves, and may trigger reactions outside the runtime's control. At the
same time, a runtime invoking a third-party tool may receive only partial
evidence of whether an effect occurred --- after a crash, a timeout, or a
lost acknowledgment. Correctness therefore depends not only on
orchestration logic, but also on semantics supplied by the lower-level
operations: outcome resolution, idempotence, compensation, staging,
dependency identity, commutativity, and visibility control.

Consider an agent booking a trip on a 1500 EUR budget. To reduce latency,
it books a 900 EUR flight and a 700 EUR hotel in parallel. Both calls
succeed, but their combined effects violate the workflow invariant. The
workflow aborts and cancels the hotel; the non-refundable flight survives.
Each operation completed correctly in isolation, yet the surviving external
state is inconsistent with the intended workflow outcome.

We argue that such failures form a common catalog with three sources, in
the anomaly-based style of classical isolation
levels~\cite{berenson1995critique,adya1999}: uncertainty about whether an
effect occurred, workflow structure that releases or depends on effects
before resolution, and interaction with concurrent executions or external
observers. The catalog is not an open-ended list: each family is enumerated
by construction. The uncertainty anomalies (A1--A3) are fixed by the three
safety-relevant actions a runtime can take under an unresolved outcome ---
re-issue, commit, or compensate; the workflow anomalies (A4--A6) by the
three ways an externalized effect can relate to its resolution point ---
survival, timing, and dependency; the interaction anomalies (A7--A8) by the
two kinds of external participant an effect can meet --- a concurrent
execution or an exogenous observer. We define correctness by excluding
anomalous patterns from an
effect history --- in the spirit of opacity~\cite{guerraoui2008opacity},
but for effects that cannot be rolled back: a safe execution is one whose
surviving external effects remain consistent with the workflow's
resolution.

The mechanisms agent runtimes use (replay, retry, compensation, staged release, coordinated access) are not new: they come from advanced
transaction models such as sagas, and from semantic recovery and
escrow-style concurrency
control~\cite{garcia1987sagas,korth1990formal,elmagarmid1992transaction,weikum2001transactional,oneil1986escrow}.
What changes in the agent setting is the boundary, where runtimes often
compose independently supplied tools through shared interfaces that expose
little of the transactional semantics those mechanisms require. Each runtime
therefore recovers the properties it needs locally (adapter annotations, registered inverses, per-effect tiers), out of band. Better replay and recovery do not close the gap on their
own: the missing piece is a contract layer at the tool interface.

This paper makes four contributions. First, an effect-history model for
agent workflows whose operations are calls to externally supplied tools
(Section~\ref{sec:model}). Second, a catalog of eight external-effect
anomalies, organized by the three ways transactional control is lost, with
guarantee profiles naming their exclusion sets and a reading of current agent
runtimes as partial coverage of the catalog (Section~\ref{sec:catalog}).
Third, a contract-level account of how the required boundary capabilities
can be exposed, together with four limits where black-box invocation alone is
insufficient, and the observation that none of the capabilities is fully
expressible in the standard MCP annotation vocabulary
(Section~\ref{sec:contracts}). Fourth, a census of how current MCP tools use
that vocabulary across reachable remote servers in the official registry
(Section~\ref{sec:grounding}).

\section{Effect Histories and the Tool Boundary}
\label{sec:model}

Agent systems form a hierarchy of levels, as in multilevel transaction
management, where the guarantees of one level are derived from the properties of
the operations below~\cite{weikum1991principles}. We work with a slice of
two adjacent levels $L_i$ and $L_{i+1}$, written L0 and L1. L0 holds the
operations the agent issues as atomic calls --- typically tools behind
external APIs, \texttt{charge()} or \texttt{book\_flight()}; L1 holds the
workflows composed from them, \texttt{BookTrip} or \texttt{HireCandidate}.
A workflow can itself serve as an operation one level up. A history spans the
slice: it records L0 attempts and external effects, the runtime's partial
observations of them, and L1 workflow resolutions. Each attempt and
compensating action, and every effect they externalize, is attributed to the
workflow and branch that issued it. Effect safety is evaluated at L1, against
those resolutions.

The slice is relative downwards too: \texttt{book\_flight} is atomic only
from the caller's side, while for its provider it is a workflow over
operations the caller does not see, of unknown depth. When that hidden
structure shows through, as when a timeout occurs inside the provider's workflow, the caller is left with an outcome it cannot confirm.

The boundary also varies in who owns it. When agent and tools are built
together, the properties of L0 operations are declared by construction, and
a runtime that owns its adapter layer is in a similar position: it can
attach the properties it needs tool by tool. Neither option transfers
cleanly across a shared interface, where a client sees only what the
interface carries. The Model Context Protocol
(MCP) is a widely used interface of this kind and is publicly inspectable.

We focus on five dimensions of L0 operations: idempotence, invertibility,
externalization timing and control, determinism, and commutativity. They are
not reducible to a single reversible/irreversible scale ---
\texttt{increment} commutes but is not idempotent, \texttt{delete} is
idempotent but does not commute. Commutativity is moreover
\emph{relational}, a property of operation pairs, and often conditional on
state (escrow-style~\cite{oneil1986escrow}); determinism, not usually
treated as an operation contract, matters because black-box LLM-backed tools
can be nondeterministic.

An execution history separates execution and external events from the
runtime's knowledge of their outcomes (Table~\ref{tab:model}). A retry is a second
attempt of the same logical operation; a compensating action externalizes in
its own right; an exogenous effect is produced by an external party rather
than directly by a runtime action; and \emph{unknown} means no
authoritative outcome is available to the runtime, although the outcome is
determined in the world. We write $externalize(q, e \text{ on } r)$ when the
resource matters. Where the individual attempt is not at issue we
write $effect(\ell)$ and $cmp(\ell)$.

\begin{table}[t]
\centering
\renewcommand{\arraystretch}{1.15}
\footnotesize
\caption{Effect-history vocabulary. The upper group records execution and
external events; the lower, runtime observations, workflow assertions, and
derived predicates.}
\label{tab:model}
\begin{tabularx}{\linewidth}{@{}l X@{}}
\toprule
$attempt(a, \ell)$ & attempt $a$ of logical operation $\ell$ \\
$externalize(q, e)$ & runtime action $q$ produced external effect $e$ \\
$externalize_{\mathit{ext}}(p, x)$ & external party $p$ produced $x$ \\
$cmp(c, a)$ & compensating action $c$ for attempt $a$ \\
$neutralizes(c, e)$ & $c$ neutralized effect $e$ \\
$dep(e_2 \leftarrow e_1)$ & $e_2$ produced having observed $e_1$ \\
$commute(e_1, e_2)$ & effects commute on their shared resource \\
$commit(w)$, $abort(w)$ & workflow $w$ is resolved \\
\midrule
$observe(a, s)$ & outcome for $a$: confirmed, failed, unknown \\
$Req(w)$ & operations that $commit(w)$ asserts took effect \\
$resolved(w)$ & $w$ has committed or aborted \\
$survives(e)$ & $e$ externalized, never neutralized \\
\bottomrule
\end{tabularx}
\end{table}

Unlike read-write histories~\cite{adya1999}, effect histories keep events
and observations apart: an effect either occurred or did not, while
\emph{unknown} is a state of the runtime, and acting under it is what
several anomalies turn on. Many of the anomalies become possible because
effects may already be irreversible or externally visible before the
workflow resolves.

\section{Effect Anomalies}
\label{sec:catalog}

Each anomaly exposes something missing at the agent--tool boundary:
evidence (A1--A3), lifecycle control (A4--A6), or coordination (A7--A8).
The uncertainty anomalies arise when the runtime cannot confirm the outcome
of an effect and must act anyway. The workflow anomalies arise from the structure of a single
execution: survival under abort, timing relative to its resolution, and
dependency. The interaction anomalies arise when an effect meets something
outside its own execution: another execution writing the same resource (A7),
or an external actor that observes and reacts to it (A8), which need not be
human --- it may itself be an agent workflow at a level our boundary does not
see.

Table~\ref{tab:catalog} states the eight patterns together with the boundary
capability needed to exclude each one in general. We derive the middle
column by asking what information or control must be available at the
boundary for a runtime protocol to rule out the anomalous history. If the
runtime cannot distinguish the anomalous history from a safe one, the
boundary must provide additional evidence. If the anomaly remains reachable
even once the relevant facts are known, exclusion additionally requires
lifecycle control or coordination. Boundary capabilities do not themselves
exclude anomalies; protocols consume them through safe retry, commit gating,
compensation, dependency tracking, and mediated ordering. The final column
gives representative ways to realize these capabilities rather than unique
or minimal implementations: some recur in the systems of
Section~\ref{sec:systems}, while others are standard transaction-processing
mechanisms or boundary primitives those systems do not expose.
Section~\ref{sec:contracts} organizes them into reusable contract families.

\begin{table}[t]
\centering
\renewcommand{\arraystretch}{1.15}
\small
\caption{The catalog of external-effect anomalies.}
\label{tab:catalog}
\begin{tabularx}{\linewidth}{@{}l X X X@{}}
\toprule
\textbf{} & \textbf{Forbidden pattern} & \textbf{Required boundary capability} & \textbf{Representative realization} \\
\midrule
A1 Duplicated & one logical operation externalizes twice & deduplication of the logical operation & logical-operation ID; idempotent re-issue \\
A2 Missing & commit without a required effect & authoritative outcome; atomic multi-effect participation & status endpoint; prepare/commit \\
A3 Orphaned & compensation issued under an unknown outcome & outcome resolution before compensating & outcome query; outcome-conditioned compensation \\
A4 Residue & aborted workflow leaves a surviving effect & residue prevention or safe neutralization & compensation declaration; staging \\
A5 Premature & possibly non-surviving effect externalizes before resolution & pre-externalization observation or control & \texttt{quote}/\texttt{dry\_run}; \texttt{hold} with expiry \\
A6 Contaminated & committed effect depends on a non-surviving effect & dependency observability and commit control & stable effect/resource identity; dependency tracking; commit gating \\
A7 Conflicting & unordered non-commuting effects from independent executions & shared-resource coordination & resource scope; commutativity declaration; mediator or ordered release \\
A8 Phantom & exogenous consequence of a compensated effect survives & control of external observability & visibility control; mediated observation \\
\bottomrule
\end{tabularx}
\end{table}

Not all of the patterns are final-state violations: A3 is defined at
action time, and A5 and A7 are preventive, relative to profiles that forbid
relying on outcomes the boundary does not establish.

\paragraph{A1: Duplicated Effect (uncertainty)}
Two attempts of the same logical operation both externalize. The
outcome of a call is not confirmed, so the runtime re-issues it, and both
attempts take effect:
\par\vspace{2pt}\noindent\begin{minipage}{\linewidth}
\begingroup\footnotesize\begin{verbatim}
attempt(a1, pay_invoice)
externalize(a1, e1)      # payment goes through
observe(a1, unknown)     # no acknowledgment
attempt(a2, pay_invoice) # re-issued
externalize(a2, e2)
=> invoice paid twice
\end{verbatim}\endgroup
\end{minipage}\par\noindent
Non-determinism makes this worse: re-issuing an operation may produce a
\emph{different} effect rather than a duplicate. An idempotent re-issue
prevents repeated externalization. Determinism does not by itself make
re-issue safe; it only makes semantic replay comparison meaningful by
making repeated executions predictable.

\paragraph{A2: Missing Committed Effect (uncertainty)}
A workflow commits relying on an effect that did not externalize.
The canonical case is uncertainty resolved optimistically: the runtime
assumes an unconfirmed effect succeeded and commits, though nothing
externalized. With $pay\_invoice \in Req(Order)$:
\par\vspace{2pt}\noindent\begin{minipage}{\linewidth}
\begingroup\footnotesize\begin{verbatim}
attempt(a1, pay_invoice)
observe(a1, unknown)     # outcome not confirmed
commit(Order)            # assumes success
=> required effect absent, yet committed
\end{verbatim}\endgroup
\end{minipage}\par\noindent
The anomaly is narrow: it is \emph{not} that the agent omitted a step (a planning failure) but that a required effect is absent from the
committed history. We focus on the uncertainty-induced case: if failure is
authoritatively known and the workflow commits anyway, the same final-state
pattern is a protocol error rather than a boundary limitation.

\paragraph{A3: Orphaned Compensation (uncertainty)}
A compensation is issued for an attempt whose outcome is unknown.
The unconfirmed outcome is resolved pessimistically instead:
\par\vspace{2pt}\noindent\begin{minipage}{\linewidth}
\begingroup\footnotesize\begin{verbatim}
attempt(a1, pay_invoice)
observe(a1, unknown)     # outcome not confirmed
cmp(c1, a1)              # compensate blindly
externalize(c1, refund)  # refund takes effect
=> if a1 never externalized, refund is spurious
\end{verbatim}\endgroup
\end{minipage}\par\noindent
The forbidden action is a compensation whose external effect does not
depend on the outcome of the original attempt. What matters is whose
knowledge decides whether an effect happens. A runtime that issues an
unconditional refund under an unknown outcome bets on the outcome, and that
bet is the violation. A compensator that checks the original operation
itself and refunds only if it occurred resolves the outcome inside the
tool, and the runtime makes no bet; its invocation is not A3 even though
the runtime's own observation is still unknown.

A1, A2, and A3 are one family: all three stem from an outcome the runtime
cannot confirm and differ only in the reaction. A1 and A2 are visible in
the final history; A3 is a violation at the moment compensation is issued:
under an unknown outcome the runtime cannot know whether compensation is
required, and even where the resulting state happens to be correct, the compensating
effect was produced without authoritative evidence. The family also marks
where this setting departs from classical transaction processing. There, an
in-doubt outcome is temporary: participants run a recovery protocol and
resolve it. A third-party tool runs no such protocol, so an unknown outcome
stays unknown until the tool itself offers a way out --- a durable
acknowledgment, a status endpoint, idempotent re-issue, or a compensator
that resolves the outcome itself.

\paragraph{A4: Uncompensated Residue (workflow)}
An aborted workflow leaves a surviving effect that was not
successfully neutralized --- no compensation exists, it was not invoked,
or it was incomplete. A non-refundable ticket is bought, then the trip is
aborted:
\par\vspace{2pt}\noindent\begin{minipage}{\linewidth}
\begingroup\footnotesize\begin{verbatim}
attempt(a1, book_flight)
externalize(a1, e1)    # non-refundable ticket
abort(BookTrip)        # budget violated
cmp(c1, a1)            # cancellation attempted
NOT neutralizes(c1, e1)
=> survives(e1); money lost
\end{verbatim}\endgroup
\end{minipage}\par

\paragraph{A5: Premature Externalization (workflow)}
An effect that may not survive is externalized before its workflow
resolves. An offer letter is sent before approval completes, and approval
is later denied:
\par\vspace{2pt}\noindent\begin{minipage}{\linewidth}
\begingroup\footnotesize\begin{verbatim}
effect(send_offer)     # letter goes out
NOT resolved(Hire)     # approval pending
abort(Hire)            # approval denied
=> send_offer preceded resolved(Hire)
\end{verbatim}\endgroup
\end{minipage}\par\noindent
A5 is profile-relative: it is not a violation under compensation-safe
execution, which may externalize an effect early and compensate later; it
is the pattern that speculation-safe execution excludes by staging or
gating effects that may not survive.

\paragraph{A6: Contaminated Speculation (workflow)}
A committed effect causally depends on an effect from a branch that
does not survive. A committed branch acts on a temporary reservation made
by a branch that is later canceled:
\par\vspace{2pt}\noindent\begin{minipage}{\linewidth}
\begingroup\footnotesize\begin{verbatim}
effect(reserve_room R) [b1]
dep(send_invites <- reserve_room R) [b2]
effect(send_invites) [b2]  # invites name R
commit(b2)                 # b2 wins
abort(b1); cmp(reserve_room R)
=> b2 announced R, which no longer survives
\end{verbatim}\endgroup
\end{minipage}\par\noindent
The contamination is causal, not about reversibility: this is the
effect-world analogue of a \emph{dirty read} (reading uncommitted state that is later rolled back), except that the ``read'' is a causal dependency on
an external effect. The harm, invitations already sent naming a room that
was released, would arise even if those invitations were themselves
reversible.

\paragraph{A7: Conflicting Externalization (interaction)}
Two independent executions release potentially non-commuting effects
on a shared resource without an ordering or mediation contract, so the
outcome may depend on their interleaving. Two coding workers publish
incompatible updates to the same shared artifact from the same base,
neither having observed the other:
\par\vspace{2pt}\noindent\begin{minipage}{\linewidth}
\begingroup\footnotesize\begin{verbatim}
effect(publish U1 on S) [w1]  # from base B
effect(publish U2 on S) [w2]  # also from B
NOT dep(U2 <- U1); NOT dep(U1 <- U2)
NOT commute(U1, U2)      # order changes result
=> outcome depends on interleaving
\end{verbatim}\endgroup
\end{minipage}\par\noindent
Unlike A6, there is no dependency between the two executions: neither
observed the other; both acted on the shared resource independently.
Commutativity \emph{excludes} the anomaly. Invertibility does not exclude
it but makes it \emph{repairable}: a versioned reversible resource lets
the loser be rolled back and re-applied toward a serializable state, as in
classical concurrency control. Mediation prevents the race by ordering
access (Section~\ref{sec:boundaries}). A7 is profile-relative, as A5 is:
the resource may serialize the calls on its own, but the externally-mediated
profile forbids relying on an outcome not established at the boundary.

\paragraph{A8: Phantom Compensation (interaction, open-world)}
An effect is compensated, but an external consequence that depends
on it survives. A supplier reacts to an offer that is later withdrawn:
\par\vspace{2pt}\noindent\begin{minipage}{\linewidth}
\begingroup\footnotesize\begin{verbatim}
externalize(a1, e)       # offer sent
externalize_ext(S, x)    # supplier acts on it
dep(x <- e)              # x saw e
cmp(c1, a1)              # offer withdrawn
neutralizes(c1, e)
survives(x)              # supplier action stands
=> offer retracted; the reaction is not
\end{verbatim}\endgroup
\end{minipage}\par\noindent
The dependent effect $x$ is exogenous: no attempt of the runtime produced it
and no compensation reaches it. This separates A8 from A6, where the
surviving effect is managed. A1--A7 concern effects issued through the managed or mediated boundary; an
exogenous consequence lies outside both the projection the runtime observes
and the set of effects it can compensate. The reacting party may
also sit outside any contractual boundary, so no ordering or commutativity
can be declared for it, and the enforceable primitive is preventive:
control whether and when the effect becomes externally observable.

\subsection{Safety Guarantee Profiles}
\label{sec:levels}

The catalog induces a vocabulary of \emph{effect-safety guarantees}, each
naming a set of anomalies it excludes: \textbf{unknown-safe} (A1--A3),
\textbf{compensation-safe} (A4, assuming compensations are correct and
succeed), \textbf{speculation-safe} (A5--A6), and
\textbf{externally-mediated} (A7--A8, where the boundary admits mediation).

Interactive tasks explain why compensation-safe is sometimes the best
achievable profile: an agent negotiating on a user's behalf cannot learn the
counterparty's reaction without sending an offer, so the action must precede
the observation and full gating is unavailable.

\subsection{Current Runtimes Against the Catalog}
\label{sec:systems}

\begin{table}[t]
\centering
\renewcommand{\arraystretch}{1.15}
\small
\caption{Coverage of the anomaly catalog by current agent runtimes.}
\label{tab:systems}
\begin{tabularx}{\linewidth}{@{}l X X@{}}
\toprule
\textbf{System} & \textbf{Targeted coverage} & \textbf{Notes} \\
\midrule
ACRFence~\cite{zheng2026acrfence} & unknown-safe partial (A1) & proposed; attack validated, not implemented \\
RAC~\cite{perera2026rac} & compensation-safe partial (A4) & no gating; no idempotency, so A1--A3 not addressed \\
Atomix~\cite{mohammadi2026atomix} & A1/A3 partial, compensation-safe (A4), speculation-safe (A5, A6), externally-mediated partial (A7) & idempotent-known-outcome path; no crash-safe exactly-once; A2 gap on multi-effect release \\
Cordon~\cite{chen2026cordon} & A1/A3 partial, compensation-safe (A4), speculation-safe partial (A5) & idempotent tools; staged effects; task-scoped \\
CoAgent~\cite{lyu2026coagent} & externally-mediated partial (A7) & reordering prevents some conflicts; registered inverses enable repair; irreversibles gated \\
Shepherd~\cite{yu2026shepherd} & per-effect reversibility tiers; A7 observed in supervisor use case & meta-agent substrate, not a runtime guarantee; irreversibles logged \\
\bottomrule
\end{tabularx}
\renewcommand{\arraystretch}{1}
\end{table}
We next use the catalog to identify which of the required capabilities
current agent runtimes target or establish under their stated assumptions.
Table~\ref{tab:systems} summarizes their
stated coverage rather than proven guarantees; none realizes a clean
cumulative hierarchy.

Atomix and Cordon stage effects, with Atomix additionally tracking
speculative dependencies; their A1/A3 coverage is conditional on assumed
tool semantics rather than on a general outcome-resolution protocol. CoAgent's reordering with registered inverses is an instance of the
achievability condition for A7 rather than an unconditional guarantee. Across these systems, coverage rests on runtime-owned declarations (adapter-declared idempotency keys, reversibility annotations, registered inverses, per-effect tiers) rather than a reusable shared contract; none controls exogenous reactions (A8), and no
system covers the full catalog. Adjacent work governs the \emph{read} side
of shared agent state~\cite{khan2026sbus} and intent revision under
conflicting irreversible actions~\cite{zhai2026revisable}, applies Saga-style validation to multi-agent planning~\cite{chang2025sagallm}, and
argues for isolation-level reasoning in workflow
systems~\cite{stonebraker2025consistency}. Agentic
Transaction~\cite{sun2026agentic} articulates ACID-oriented semantic
guarantees for agent systems and instantiates them in a runtime for data
agents; we ask instead what the boundary to third-party tools must declare
for guarantees of this kind to be establishable at all. A recent formal
catalog of concurrency anomalies for multi-agent runtimes comes with
mechanically verified detectors and safety
guarantees~\cite{khan2026verified}; its causal-cascade anomaly, in which a
dependent operation survives an aborted ancestor, is close to our A6, and
the model likewise admits external effects that no internal rollback can
undo. Our catalog differs in what it centers: the outcome uncertainty of
calls to independently supplied tools, and the interface capabilities a
boundary must expose to exclude each anomaly, rather than verified
protocols over a runtime's shared state.

\subsection{Scope and Coverage}
\label{sec:coverage}

The catalog is coverage-oriented rather than complete by theorem: absolute
completeness is unavailable in an open world, where an external actor can
react in unbounded ways. Each family has a structural reason for its
membership. The uncertainty family is fixed by the safety-relevant actions a runtime
can take while an outcome is unresolved: re-issue, commit as though it had
externalized, or compensate. Waiting, querying, or escalating creates no new
effect-history pattern until one of these is taken. The workflow family is fixed by how an
externalized effect relates to its resolution point --- survival, timing,
dependency. The interaction family separates unmediated concurrent effects
on a shared resource from an exogenous consequence of a later-compensated
effect. We conjecture that, relative to the vocabulary of
Section~\ref{sec:model}, every loss of transactional control the model can
represent falls into one of the three families; formal coverage,
minimality, and independence are future work.

We place the following outside the model, so the catalog is not mistaken for covering them: semantic planning errors (the agent booked the wrong city); read-side
anomalies; security and policy
violations; tool-contract misclassification (a tool declared reversible that
is not); intra-execution effect order (formalized for multi-agent runtimes
by~\cite{khan2026verified}), except insofar as it participates in an A7
conflict; and liveness failures.

\section{Transactional Tool Contracts and Guarantee Boundaries}
\label{sec:contracts}
\label{sec:unknown}

Advanced transaction models assume properties such as idempotence,
invertibility, and commutativity are known; at a third-party boundary they
are often absent or inferred. A reusable contract layer must
therefore distinguish established, assumed, and unknown properties, since
acting on an assumption, such as compensating an operation with no true inverse, is a failure mode of its own. Such declarations do not replace runtime
protocols: they are their precondition.

A runtime consuming tools through a shared interface reads only what that
interface carries, and MCP is not entirely silent: since its 2025-03-26
revision, tool annotations let a server flag a tool as read-only,
destructive, idempotent, or open-world~\cite{mcp2025spec}. These are advisory boolean hints ---
unenforced, optional, and coarser than the capabilities of
Table~\ref{tab:catalog}: \texttt{idempotentHint} asserts that repetition is
safe but supplies no idempotency key, so a tool that accepts one can declare
the fact only in its input schema, outside the annotation vocabulary; and no
hint covers status resolution, compensation, staging, commutativity, or
visibility. The 2025-11-25 revision added experimental
tasks~\cite{mcp2025tasks}, which return a durable handle that the caller
polls for a terminal status. This supplies part of the observability A1
and A2 need, but nothing of the rest: tasks carry no idempotency,
compensation, or commit semantics, cancellation is cooperative rather
than an undo, and the handle can itself be lost before it is received,
which moves an unknown outcome one request earlier rather than removing
it.
Table~\ref{tab:gap} goes through the capabilities one by one. None of these
capabilities is fully expressible in the annotation vocabulary, so the standard annotations alone
provide no reusable transactional contract. A tool can still arrange such
a contract on top of MCP, through its input schema, its description, or
tasks, but that contract is particular to the tool and must be learned tool
by tool. This is a property of the specification. How the four fields it
does define are used in practice is the subject of
Section~\ref{sec:grounding}.

\begin{table}[t]
\centering
\renewcommand{\arraystretch}{1.1}
\footnotesize
\caption{Expressibility of the boundary capabilities in current MCP tool
annotations.}
\label{tab:gap}
\begin{tabularx}{\linewidth}{@{}X l@{}}
\toprule
\textbf{Boundary capability} & \textbf{In annotations} \\
\midrule
A1 deduplication of the logical operation & Limited: idempotence hint only \\
A2 authoritative outcome; atomic participation & No \\
A3 outcome resolution before compensating & No \\
A4 residue prevention or neutralization & No \\
A5 pre-externalization observation/control & No \\
A6 dependency observability; commit control & No \\
A7 shared-resource coordination & No: externality only \\
A8 control of external observability & No \\
\bottomrule
\end{tabularx}
\end{table}

\subsection{Recurring Contract Families}

The capabilities in Table~\ref{tab:catalog} fall into three contract
families. A capability is what the boundary lets a runtime establish; a
contract is its reusable declaration, possibly spanning several
operations.

Uncertainty requires authoritative convergence on one logical outcome.
Deduplication alone excludes A1 but does not resolve the
outcome that A2 and A3 turn on. The adapter-declared idempotency key is the
closest primitive in use among the systems of Section~\ref{sec:systems};
atomic multi-effect release is deferred to
Section~\ref{sec:boundaries}.

Lifecycle control (A4--A6) needs the workflow's invariants to be checkable
before irreversible effects occur, compensation semantics for what a
compensator requires and what counts as its success, and stable effect and
resource identity afterwards. The first of these is a condition on the
boundary rather than on the runtime, since gating is available only where
the values an invariant depends on can be observed without an effect. A
read-only \texttt{quote} provides such an observation and an expiring
\texttt{hold} replaces the final effect with a provisional one whose
visibility, expiry, and release semantics are declared, while a
negotiation-style interaction offers neither. The unit dependency
tracking protects is a \emph{commit sphere}, the minimal set of effects that
must become final together.

Coordination (A7--A8) requires contracts at a different granularity.
Idempotency, status, quote, and compensation are properties of a single
operation, whereas commutativity holds of a \emph{pair} of operations on a
resource and cannot be set by rules given to one agent: it must be declared
at the resource; where it does not hold, exclusion requires a mediator that
serializes access, and declared inverses support repair after a conflict. Visibility is likewise a property of the boundary:
who may observe an effect before it is final.

A boundary that exposes these capabilities makes the corresponding
exclusions implementable by a conforming runtime, given a suitable protocol
and declarations that hold, rather than runtime-local out-of-band
knowledge.

\subsection{Guarantee Boundaries Above Black-Box Tools}
\label{sec:boundaries}

Four boundaries mark where black-box invocation alone is insufficient:
stronger guarantees require authoritative tool participation, shared
mediation, or visibility control. We state them as consequences of the
definitions and relate them to current systems; formal proofs are future
work.

First, without a primitive that establishes one authoritative outcome (idempotent re-issue returning the original outcome, a status endpoint, or a durable acknowledgment), no protocol can both resolve the workflow after
an ambiguous tool call and guarantee unknown-safe and compensation-safe
execution. Over an unreliable channel the runtime cannot learn in bounded
time whether an effect took place, the classical exactly-once
barrier~\cite{bernstein1987concurrency} at the agent--tool boundary.
Re-issuing risks A1, committing risks A2, compensating risks A3, and
aborting without compensation risks A4 if the effect occurred. Waiting
avoids this resolution-time choice but gives up resolution.

Second, non-commuting irreversible effects without a mediator admit no
general conflict repair. Without operation-specific reconciliation, the
runtime has no general way to transform the realized state into one
corresponding to a serial order; undo and re-application would provide such
a path, but irreversibility rules it out. What
remains is prevention: mediation before unordered externalization, or gating
conflicting calls until earlier ones commit, as
CoAgent~\cite{lyu2026coagent} does.

Third, an open-world reaction ends the reach of compensation: once an outside
actor has observed a released effect and acted, retracting the effect does
not retract the reaction (A8). The available contract is preventive ---
visibility control before observation.

Fourth, several irreversible effects on different tools cannot be released
atomically above the tool layer. A crash or persistent failure between
releases leaves a partial externalization that the runtime cannot generally
complete or undo: committing risks a missing required effect (A2), while
aborting leaves surviving residue (A4). Avoiding this choice is the
classical atomic-commitment problem~\cite{bernstein1987concurrency} and
needs tool-side prepare/commit participation, a limit
Atomix~\cite{mohammadi2026atomix} states for its own release step.

\section{What the Boundary Declares Today}
\label{sec:grounding}

The vocabulary cannot carry the capabilities of Section~\ref{sec:contracts},
but it does carry four fields, and what servers put in them decides what
a runtime can learn today. We measured this with a census of 98{,}291
tools exposed by reachable remote servers in the official MCP registry
(snapshot 2026-07-27), recording
\texttt{readOnlyHint}, \texttt{destructiveHint}, \texttt{idempotentHint},
and \texttt{openWorldHint} for each tool and keeping an explicit
\texttt{false} distinct from an omitted field.

\paragraph{Sampling and methodology.}
We took a full snapshot of the official MCP registry on 2026-07-27
(59{,}625 entries; 18{,}688 distinct servers, keeping the latest version of
each). Of these, 9{,}454 exposed no remote endpoint and were out of scope
because probing them would require executing a package or stdio server. We
anonymously queried \texttt{tools/list} on all 9{,}234 remote targets,
including those declaring an authentication header; no responding target
rejected discovery with HTTP 401 or 403 (connection failures were not
classified). Of these targets, 4{,}838 returned
at least one tool, 4{,}318 failed to connect, 74 timed out, and four
returned no tools, yielding 98{,}291 tools (median 11 per server). No
tool was called. The resulting measurements describe the reachable remote
subset; unreachable and package/stdio-only servers may differ. We do not attribute measured values to author intent. Code, data,
and full methodology are available in the accompanying
artifact.\footnote{\url{https://github.com/flame-stream/mcp-annotation-census}}

\paragraph{Annotation fields are widely emitted, by one publisher.}
At the wire level, 74.0\% of tools serialize at least one field and 61.7\%
serialize all four. Most of this is one fleet. A single publisher,
\texttt{io.github.pipeworx-io}, contributes 1{,}274 of the 4{,}838
responding servers and 44{,}227 tools, 45.0\% of the corpus, every one of
them with all four fields set. Without it, 52.8\% of the remaining tools
serialize at least one field and 30.3\% serialize all four. A field being
present therefore says little about whether it was chosen, and some fields
carry no information even when set: \texttt{destructiveHint} is
inapplicable when \texttt{readOnlyHint} is true, a pairing that accounts for
52.8\% of all tools and 26.0\% without the dominant publisher.

\paragraph{Tool-level signatures are concentrated.}
Reading each tool as a signature over the four fields (each true, false, or
omitted), 66 of the 81 possible signatures appear, but one signature
(read-only, non-destructive, idempotent, open-world) covers 39.9\% of tools
and the next most common is no annotation at all (26.0\%); the top three
together cover 75.8\%. The dominant signature is the dominant publisher's
template: without that publisher it falls to 7.3\% and no annotation at all
becomes the most common signature, at 47.2\% of tools. Among multi-tool
servers that emit at least one field, 76.7\% use more than one signature,
and across emitting servers the median dominant signature covers 79.4\% of
a server's tools: differentiation is widespread but coarse.

\paragraph{What the hints do not resolve.}
Even fully populated, the four fields describe the shape of a call while leaving its transactional semantics unstated: none gives an idempotency key, a status endpoint, a compensation contract, or a commutativity rule. Take \texttt{destructiveHint}, the one field about risk. It is set on
65.8\% of all tools, but is meaningful only where the tool is not
read-only: just 12.9\% of tools carry an applicable classification, and
only 3.1\% assert an actual destructive operation. Unlike the emission
figures, these two do not depend on the dominant publisher: without it
they are 11.7\% and 3.2\%.

\medskip

\noindent
The census reads
the boundary as declared, not as implementations behave. Tools may enforce
stronger semantics internally (deduplication, idempotency keys) than any
annotation surfaces; a runtime that relies on the annotations cannot use that
safety, and measuring the gap requires implementation-level
analysis~\cite{toeppe2026mcpdataset}.

\section{Conclusion and Future Work}

We introduced an effect-history model, a catalog of eight external-effect
anomalies with the boundary capability that excludes each, and four limits
of black-box invocation. Current runtimes cover fragments of the catalog,
and the standard MCP annotations express none of the capabilities in full:
the missing layer is reusable transactional contracts at the tool boundary.
Future work includes formal proofs of the limits, protocol synthesis from
declared contracts, and a cost model that prices anomalies against their
compensations.

\setlength{\bibsep}{0pt plus 0.5pt}
\bibliographystyle{unsrtnat}
\bibliography{bibliography/flame-stream}

\end{document}